# Exploiting Causal Independence in Bayesian Network Inference

**Nevin Lianwen Zhang**                                        LZHANG@CS.UST.HK
*Department of Computer Science,*
*University of Science & Technology, Hong Kong*

**David Poole**                                                POOLE@CS.UBC.CA
*Department of Computer Science, University of British Columbia,*
*2366 Main Mall, Vancouver, B.C., Canada V6T 1Z4*

## Abstract

A new method is proposed for exploiting causal independencies in exact Bayesian network inference. A Bayesian network can be viewed as representing a factorization of a joint probability into the multiplication of a set of conditional probabilities. We present a notion of causal independence that enables one to further factorize the conditional probabilities into a combination of even smaller factors and consequently obtain a finer-grain factorization of the joint probability. The new formulation of causal independence lets us specify the conditional probability of a variable given its parents in terms of an associative and commutative operator, such as "or", "sum" or "max", on the contribution of each parent. We start with a simple algorithm VE for Bayesian network inference that, given evidence and a query variable, uses the factorization to find the posterior distribution of the query. We show how this algorithm can be extended to exploit causal independence. Empirical studies, based on the CPCS networks for medical diagnosis, show that this method is more efficient than previous methods and allows for inference in larger networks than previous algorithms.

## 1. Introduction

Reasoning with uncertain knowledge and beliefs has long been recognized as an important research issue in AI (Shortliffe & Buchanan, 1975; Duda et al., 1976). Several methodologies have been proposed, including certainty factors, fuzzy sets, Dempster-Shafer theory, and probability theory. The probabilistic approach is now by far the most popular among all those alternatives, mainly due to a knowledge representation framework called Bayesian networks or belief networks (Pearl, 1988; Howard & Matheson, 1981).

Bayesian networks are a graphical representation of (in)dependencies amongst random variables. A Bayesian network (BN) is a DAG with nodes representing random variables, and arcs representing direct influence. The independence that is encoded in a Bayesian network is that each variable is independent of its non-descendents given its parents.

Bayesian networks aid in knowledge acquisition by specifying which probabilities are needed. Where the network structure is sparse, the number of probabilities required can be much less than the number required if there were no independencies. The structure can be exploited computationally to make inference faster (Pearl, 1988; Lauritzen & Spiegelhalter, 1988; Jensen et al., 1990; Shafer & Shenoy, 1990).

The definition of a Bayesian network does not constrain how a variable depends on its parents. Often, however, there is much structure in these probability functions that can be exploited for knowledge acquisition and inference. One such case is where some dependencies depend on particular values of other variables; such dependencies can be stated as rules (Poole, 1993), trees (Boutilier et al., 1996) or as multinets (Geiger & Heckerman, 1996). Another is where the the function can be described using a binary operator that can be applied to values from each of the parent variables. It is the latter, known as 'causal independencies', that we seek to exploit in this paper.

Causal independence refers to the situation where multiple causes contribute independently to a common effect. A well-known example is the noisy OR-gate model (Good, 1961). Knowledge engineers have been using specific causal independence models in simplifying knowledge acquisition (Henrion, 1987; Olesen et al., 1989; Olesen & Andreassen, 1993). Heckerman (1993) was the first to formalize the general concept of causal independence. The formalization was later refined by Heckerman and Breese (1994).

Kim and Pearl (1983) showed how the use of noisy OR-gate can speed up inference in a special kind of BNs known as polytrees; D'Ambrosio (1994, 1995) showed the same for two level BNs with binary variables. For general BNs, Olesen *et al.* (1989) and Heckerman (1993) proposed two ways of using causal independencies to transform the network structures. Inference in the transformed networks is more efficient than in the original networks (see Section 9).

This paper proposes a new method for exploiting a special type of causal independence (see Section 4) that still covers common causal independence models such as noisy OR-gates, noisy MAX-gates, noisy AND-gates, and noisy adders as special cases. The method is based on the following observation. A BN can be viewed as representing a factorization of a joint probability into the multiplication of a list of conditional probabilities (Shachter et al., 1990; Zhang & Poole, 1994; Li & D'Ambrosio, 1994). The type of causal independence studied in this paper leads to further factorization of the conditional probabilities (Section 5). A finer-grain factorization of the joint probability is obtained as a result. We propose to extend exact inference algorithms that only exploit conditional independencies to also make use of the finer-grain factorization provided by causal independence.

The state-of-art exact inference algorithm is called clique tree propagation (CTP) (Lauritzen & Spiegelhalter, 1988; Jensen et al., 1990; Shafer & Shenoy, 1990). This paper proposes another algorithm called variable elimination (VE) (Section 3), that is related to SPI (Shachter et al., 1990; Li & D'Ambrosio, 1994), and extends it to make use of the finer-grain factorization (see Sections 6, 7, and 8). Rather than compiling to a secondary structure and finding the posterior probability for each variable, VE is query-oriented; it needs only that part of the network relevant to the query given the observations, and only does the work necessary to answer that query. We chose VE instead of CTP because of its simplicity and because it can carry out inference in large networks that CTP cannot deal with.

Experiments (Section 10) have been performed with two CPCS networks provided by Pradhan. The networks consist of 364 and 421 nodes respectively and they both contain abundant causal independencies. Before this paper, the best one could do in terms of exact inference would be to first transform the networks by using Jensen *et al.*'s or Heckerman's technique and then apply CTP. In our experiments, the computer ran out of memory when constructing clique trees for the transformed networks. When this occurs one cannot answer any query at all. However, the extended VE algorithm has been able to answer almost all randomly generated queries with twenty or less observations (findings) in both networks.

One might propose to first perform Jensen *et al.*'s or Heckerman's transformation and then apply VE. Our experiments show that this is significantly less efficient than the extended VE algorithm.

We begin with a brief review of the concept of a Bayesian network and the issue of inference.

## 2. Bayesian Networks

We assume that a problem domain is characterized by a set of random variables. Beliefs are represented by a *Bayesian network* (BN) — an annotated directed acyclic graph, where nodes represent the random variables, and arcs represent probabilistic dependencies amongst the variables. We use the terms 'node' and 'variable' interchangeably. Associated with each node is a conditional probability of the variable given its parents.

In addition to the explicitly represented conditional probabilities, a BN also implicitly represents conditional independence assertions. Let $x_1, x_2, ..., x_n$ be an enumeration of all the nodes in a BN such that each node appears after its children, and let $\pi_{x_i}$ be the set of parents of a node $x_i$. The Bayesian network represents the following independence assertion:

> Each variable $x_i$ is conditionally independent of the variables in $\{x_1, x_2, \ldots, x_{i-1}\}$ given values for its parents.

The conditional independence assertions and the conditional probabilities together entail a joint probability over all the variables. By the chain rule, we have:

$$
\begin{aligned}
P(x_1, x_2, \ldots, x_n) &= \prod_{i=1}^{n} P(x_i | x_1, x_2, \ldots, x_{i-1}) \\
&= \prod_{i=1}^{n} P(x_i | \pi_{x_i}),
\end{aligned}
\tag{1}
$$

where the second equation is true because of the conditional independence assertions. The conditional probabilities $P(x_i | \pi_{x_i})$ are given in the specification of the BN. Consequently, one can, in theory, do arbitrary probabilistic reasoning in a BN.

### 2.1 Inference

*Inference* refers to the process of computing the posterior probability $P(X | Y = Y_0)$ of a set $X$ of query variables after obtaining some observations $Y = Y_0$. Here $Y$ is a list of observed variables and $Y_0$ is the corresponding list of observed values. Often, $X$ consists of only one query variable.

In theory, $P(X | Y = Y_0)$ can be obtained from the marginal probability $P(X, Y)$, which in turn can be computed from the joint probability $P(x_1, x_2, \ldots, x_n)$ by summing out variables outside $X \cup Y$ one by one. In practice, this is not viable because summing out a variable from a joint probability requires an exponential number of additions.

The key to more efficient inference lies in the concept of factorization. A *factorization* of a joint probability is a list of *factors* (functions) from which one can construct the joint probability.

A *factor* is a function from a set of variables into a number. We say that the factor *contains* a variable if the factor is a function of that variable; or say it is a factor *of* the variables on which it depends. Suppose $f_1$ and $f_2$ are factors, where $f_1$ is a factor that contains variables $x_1, \ldots, x_i, y_1, \ldots, y_j$ — we write this as $f_1(x_1, \ldots, x_i, y_1, \ldots, y_j)$ — and $f_2$ is a factor with variables $y_1, \ldots, y_j, z_1, \ldots, z_k$, where $y_1, \ldots, y_j$ are the variables in common to $f_1$ and $f_2$. The product of $f_1$ and $f_2$ is a factor that is a function of the union of the variables, namely $x_1, \ldots, x_i, y_1, \ldots, y_j, z_1, \ldots, z_k$, defined by:

$$(f_1 \times f_2)(x_1, \ldots, x_i, y_1, \ldots, y_j, z_1, \ldots, z_k) = f_1(x_1, \ldots, x_i, y_1, \ldots, y_j) \times f_2(y_1, \ldots, y_j, z_1, \ldots, z_k)$$

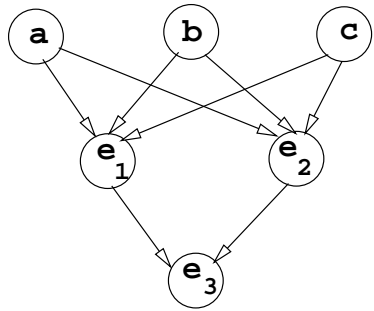

Figure 1: A Bayesian network.

Let $f(x_1, \ldots, x_i)$ be a function of variable $x_1, \ldots, x_i$. *Setting*, say $x_1$ in $f(x_1, \ldots, x_i)$ to a particular value $\alpha$ yields $f(x_1=\alpha, x_2, \ldots, x_i)$, which is a function of variables $x_2, \ldots, x_i$.

If $f(x_1, \ldots, x_i)$ is a factor, we can sum out a variable, say $x_1$, resulting in a factor of variables $x_2, \ldots, x_i$, defined

$$(\sum_{x_1} f)(x_2, \ldots, x_i) = f(x_1=\alpha_1, x_2, \ldots, x_i) + \cdots + f(x_1=\alpha_m, x_2, \ldots, x_i)$$

where $\alpha_1, \ldots, \alpha_m$ are the possible values of variable $x_1$.

Because of equation (1), a BN can be viewed as representing a factorization of a joint probability. For example, the Bayesian network in Figure 1 factorizes the joint probability $P(a, b, c, e_1, e_2, e_3)$ into the following list of factors:

$$P(a), P(b), P(c), P(e_1|a, b, c), P(e_2|a, b, c), P(e_3|e_1, e_2).$$

Multiplying those factors yields the joint probability.

Suppose a joint probability $P(z_1, z_2, \ldots, z_m)$ is factorized into the multiplication of a list of factors $f_1, f_2, \ldots, f_m$. While obtaining $P(z_2, \ldots, z_m)$ by summing out $z_1$ from $P(z_1, z_2, \ldots, z_m)$ requires an exponential number of additions, obtaining a factorization of $P(z_2, \ldots, z_m)$ can often be done with much less computation. Consider the following procedure:

Procedure `sum-out`$(\mathcal{F}, z)$:

- Inputs: $\mathcal{F}$ — a list of factors; $z$ — a variable.

- Output: A list of factors.

1. Remove from the $\mathcal{F}$ all the factors, say $f_1, \ldots, f_k$, that contain $z$,

2. Add the new factor $\sum_z \prod_{i=1}^k f_i$ to $\mathcal{F}$ and return $\mathcal{F}$.

**Theorem 1** *Suppose a joint probability $P(z_1, z_2, \ldots, z_m)$ is factorized into the multiplication of a list $\mathcal{F}$ of factors. Then* `sum-out`$(\mathcal{F}, z_1)$ *returns a list of factors whose multiplication is $P(z_2, \ldots, z_m)$.*

**Proof**: Suppose $\mathcal{F}$ consists of factors $f_1, f_2, \ldots, f_m$ and suppose $z_1$ appears in and only in factors $f_1, f_2, \ldots, f_k$. Then

$$
\begin{aligned}
P(z_2, \ldots, z_m) &= \sum_{z_1} P(z_1, z_2, \ldots, z_m) \\
&= \sum_{z_1} \prod_{i=1}^{m} f_i = [\sum_{z_1} \prod_{i=1}^{k} f_i][\prod_{i=k+1}^{m} f_i].
\end{aligned}
$$

The theorem follows. $\square$

Only variables that appear in the factors $f_1, f_2, \ldots, f_k$ participated in the computation of sum-out$(\mathcal{F}, z_1)$, and those are often only a small portion of all the variables. This is why inference in a BN can be tractable in many cases, even if the general problem is NP-hard (Cooper, 1990).

## 3. The Variable Elimination Algorithm

Based on the discussions of the previous section, we present a simple algorithm for computing $P(X|Y=Y_0)$. The algorithm is based on the intuitions underlying D'Ambrosio's symbolic probabilistic inference (SPI) (Shachter et al., 1990; Li & D'Ambrosio, 1994), and first appeared in Zhang and Poole (1994). It is essentially Dechter (1996)'s bucket elimination algorithm for belief assessment.

The algorithm is called variable elimination (VE) because it sums out variables from a list of factors one by one. An ordering $\rho$ by which variables outside $X \cup Y$ to be summed out is required as an input. It is called an *elimination ordering*.

Procedure VE$(\mathcal{F}, X, Y, Y_0, \rho)$

- Inputs: $\mathcal{F}$ — The list of conditional probabilities in a BN;
  $X$ — A list of query variables;
  $Y$ — A list of observed variables;
  $Y_0$ — The corresponding list of observed values;
  $\rho$ — An elimination ordering for variables outside $X \cup Y$.

- Output: $P(X|Y=Y_0)$.

1. Set the observed variables in all factors to their corresponding observed values.

2. **While** $\rho$ is not empty,

   (a) Remove the first variable $z$ from $\rho$,

   (b) Call sum-out$(\mathcal{F}, z)$. **Endwhile**

3. Set $h =$ the multiplication of all the factors on $\mathcal{F}$.
   /* h is a function of variables in $X$. */

4. Return $h(X) / \sum_X h(X)$. /* Renormalization */

**Theorem 2** *The output of* VE$(\mathcal{F}, X, Y, Y_0, \rho)$ *is indeed* $P(X|Y=Y_0)$.

**Proof**: Consider the following modifications to the procedure. First remove step 1. Then the factor $h$ produced at step 3 will be a function of variables in $X$ and $Y$. Add a new step after step 3 that sets the observed variables in $h$ to their observed values.

Let $f(y, A)$ be a function of variable $y$ and of variables in $A$. We use $f(y, A)|_{y=\alpha}$ to denote $f(y=\alpha, A)$. Let $f(y, -)$, $g(y, -)$, and $h(y, z, -)$ be three functions of $y$ and other variables. It is evident that

$$f(y, -)g(y, -)|_{y=\alpha} = f(y, -)|_{y=\alpha} \, g(y, -)|_{y=\alpha},$$

$$[\sum_z h(y, z, -)]|_{y=\alpha} = \sum_z [h(y, z, -)|_{y=\alpha}].$$

Consequently, the modifications do not change the output of the procedure.

According to Theorem 1, after the modifications the factor produced at step 3 is simply the marginal probability $P(X, Y)$. Consequently, the output is exactly $P(X|Y=Y_0)$. □

The complexity of VE can be measured by the number of numerical multiplications and numerical summations it performs. An *optimal elimination ordering* is one that results in the least complexity. The problem of finding an optimal elimination ordering is NP-complete (Arnborg et al., 1987). Commonly used heuristics include minimum deficiency search (Bertelè & Brioschi, 1972) and maximum cardinality search (Tarjan & Yannakakis, 1984). Kjærulff (1990) has empirically shown that minimum deficiency search is the best existing heuristic. We use minimum deficiency search in our experiments because we also found it to be better than the maximum cardinality search.

### 3.1 VE versus Clique Tree Propagation

Clique tree propagation (Lauritzen & Spiegelhalter, 1988; Jensen et al., 1990; Shafer & Shenoy, 1990) has a compilation step that transforms a BN into a secondary structure called clique tree or junction tree. The secondary structure allows CTP to compute the answers to all queries with one query variable and a fixed set of observations in twice the time needed to answer one such query in the clique tree. For many applications this is a desirable property since a user might want to compare the posterior probabilities of different variables.

CTP takes work to build the secondary structure before any observations have been received. When the Bayesian network is reused, the cost of building the secondary structure can be amortized over many cases. Each observation entails a propagation though the network.

Given all of the observations, VE processes one query at a time. If a user wants the posterior probabilities of several variables, or for a sequence of observations, she needs to run VE for each of the variables and observation sets.

The cost, in terms of the number of summations and multiplications, of answering a single query with no observations using VE is of the same order of magnitude as using CTP. A particular clique tree and propagation sequence encodes an elimination ordering; using VE on that elimination ordering results in approximately the same summations and multiplications of factors as in the CTP (there is some discrepancy, as VE does not actually form the marginals on the cliques, but works with conditional probabilities directly). Observations make VE simpler (the observed variables are eliminated at the start of the algorithm), but each observation in CTP requires propagation of evidence. Because VE is query oriented, we can prune nodes that are irrelevant to specific queries (Geiger et al., 1990; Lauritzen et al., 1990; Baker & Boult, 1990). In CTP, on the other hand, the clique tree structure is kept static at run time, and hence does not allow pruning of irrelevant nodes.

CTP encodes a particular space-time tradeoff, and VE another. CTP is particularly suited to the case where observations arrive incrementally, where we want the posterior probability of each node, and where the cost of building the clique tree can be amortized over many cases. VE is suited for one-off queries, where there is a single query variable and all of the observations are given at once.

Unfortunately, there are large real-world networks that CTP cannot deal with due to time and space complexities (see Section 10 for two examples). In such networks, VE can still answer some of the possible queries because it permits pruning of irrelevant variables.

## 4. Causal Independence

Bayesian networks place no restriction on how a node depends on its parents. Unfortunately this means that in the most general case we need to specify an exponential (in the number of parents) number of conditional probabilities for each node. There are many cases where there is structure in the probability tables that can be exploited for both acquisition and for inference. One such case that we investigate in this paper is known as 'causal independence'.

In one interpretation, arcs in a BN represent causal relationships; the parents $c_1, c_2, \ldots, c_m$ of a variable $e$ are viewed as causes that jointly bear on the effect $e$. Causal independence refers to the situation where the causes $c_1, c_2, \ldots, c_m$ contribute independently to the effect $e$.

More precisely, $c_1, c_2, \ldots, c_m$ are said to be *causally independent* w.r.t. effect $e$ if there exist random variables $\xi_1, \xi_2, \ldots, \xi_m$ that have the same *frame*, i.e., the same set of possible values, as $e$ such that

1. For each $i$, $\xi_i$ probabilistically depends on $c_i$ and is conditionally independent of all other $c_j$'s and all other $\xi_j$'s given $c_i$, and

2. There exists a commutative and associative binary operator $*$ over the frame of $e$ such that $e = \xi_1 * \xi_2 * \ldots * \xi_m$.

Using the independence notion of Pearl (1988), let $I(X, Y|Z)$ mean that $X$ is independent of $Y$ given $Z$, the first condition is:

$$I(\xi_1, \{c_2, \ldots, c_m, \xi_2, \ldots, \xi_m\}|c_1)$$

and similarly for the other variables. This entails $I(\xi_1, c_j|c_1)$ and $I(\xi_1, \xi_j|c_1)$ for each $c_j$ and $\xi_j$ where $j \neq 1$.

We refer to $\xi_i$ as the *contribution* of $c_i$ to $e$. In less technical terms, causes are causally independent w.r.t. their common effect if individual contributions from different causes are independent and the total influence on the effect is a combination of the individual contributions.

We call the variable $e$ a *convergent variable* as it is where independent contributions from different sources are collected and combined (and for the lack of a better name). Non-convergent variables will simply be called *regular variables*. We call $*$ the *base combination operator* of $e$.

The definition of causal independence given here is slightly different than that given by Heckerman and Breese (1994) and Srinivas (1993). However, it still covers common causal independence models such as noisy OR-gates (Good, 1961; Pearl, 1988), noisy MAX-gates (Díez, 1993), noisy AND-gates, and noisy adders (Dagum & Galper, 1993) as special cases. One can see this in the following examples.

**Example 1 (Lottery)** Buying lotteries affects your wealth. The amounts of money you spend on buying different kinds of lotteries affect your wealth independently. In other words, they are causally independent w.r.t. the change in your wealth. Let $c_1, \ldots, c_k$ denote the amounts of money you spend on buying $k$ types of lottery tickets. Let $\xi_1, \ldots, \xi_k$ be the changes in your wealth due to buying the different types of lottery tickets respectively. Then, each $\xi_i$ depends probabilistically on $c_i$ and is conditionally independent of the other $c_j$ and $\xi_j$ given $c_i$. Let $\epsilon$ be the total change in your wealth due to lottery buying. Then $\epsilon = \xi_1 + \cdots + \xi_k$. Hence $c_1, \ldots, c_k$ are causally independent w.r.t. $\epsilon$. The base combination operator of $\epsilon$ is numerical addition. This example is an instance of a causal independence model called *noisy adders*.

If $c_1, \ldots, c_k$ are the amounts of money you spend on buying lottery tickets in the same lottery, then $c_1, \ldots, c_k$ are not causally independent w.r.t. $\epsilon$, because winning with one ticket reduces the chance of winning with the other. Thus, $\xi_1$ is not conditionally independent of $\xi_2$ given $c_1$. However, if the $c_i$ represent the expected change in wealth in buying tickets in the same lottery, then they would be causally independent, but not probabilistically independent (there would be arcs between the $c_i$'s).

**Example 2 (Alarm)** Consider the following scenario. There are $m$ different motion sensors each of which are connected to a burglary alarm. If one sensor activates, then the alarm rings. Different sensors could have different reliability. We can treat the activation of sensor $i$ as a random variable. The reliability of the sensor can be reflected in the $\xi_i$. We assume that the sensors fail independently[1]. Assume that the alarm can only be caused by a sensor activation[2]. Then $alarm = \xi_1 \vee \cdots \vee \xi_m$; the base combination operator here is the logical OR operator. This example is an instance of a causal independence model called the *noisy OR-gate*.

The following example is not an instance of any causal independence models that we know:

**Example 3 (Contract renewal)** Faculty members at a university are evaluated in teaching, research, and service for the purpose of contract renewal. A faculty member's contract is not renewed, renewed without pay raise, renewed with a pay raise, or renewed with double pay raise depending on whether his performance is evaluated unacceptable in at least one of the three areas, acceptable in all areas, excellent in one area, or excellent in at least two areas.

Let $c_1$, $c_2$, and $c_3$ be the fractions of time a faculty member spends on teaching, research, and service respectively. Let $\xi_i$ represent the evaluation he gets in the $i$th area. It can take values 0, 1, and 2 depending on whether the evaluation is unacceptable, acceptable, or excellent. The variable $\xi_i$ depends probabilistically on $c_i$. It is reasonable to assume that $\xi_i$ is conditionally independent of other $c_j$'s and other $\xi_j$'s given $c_i$.

Let $\epsilon$ represent the contract renewal result. The variable can take values 0, 1, 2, and 3 depending on whether the contract is not renewed, renewed with no pay raise, renewed with a pay raise, or renewed with double pay raise. Then $\epsilon = \xi_1 * \xi_2 * \xi_3$, where the base combination operator $*$ is given in this following table:

|   | 0 | 1 | 2 | 3 |
|---|---|---|---|---|
| 0 | 0 | 0 | 0 | 0 |
| 1 | 0 | 1 | 2 | 3 |
| 2 | 0 | 2 | 3 | 3 |
| 3 | 0 | 3 | 3 | 3 |

---

1. This is called the *exception independence assumption* by Pearl (1988).

2. This is called the *accountability assumption* by Pearl (1988). The assumption can always be satisfied by introducing a node that represent all other causes (Henrion, 1987).

So, the fractions of time a faculty member spends in the three areas are causally independent w.r.t. the contract renewal result.

In the traditional formulation of a Bayesian network we need to specify an exponential, in the number of parents, number of conditional probabilities for a variable. With causal independence, the number of conditional probabilities $P(\xi_i|c_i)$ is linear in $m$. This is why causal independence can reduce complexity of knowledge acquisition (Henrion, 1987; Pearl, 1988; Olesen et al., 1989; Olesen & Andreassen, 1993). In the following sections we show how causal independence can also be exploited for computational gain.

### 4.1 Conditional Probabilities of Convergent Variables

VE allows us to exploit structure in a Bayesian network by providing a factorization of the joint probability distribution. In this section we show how causal independence can be used to factorize the joint distribution even further. The initial factors in the VE algorithm are of the form $P(e|c_1, \ldots, c_m)$. We want to break this down into simpler factors so that we do not need a table exponential in $m$. The following proposition shows how causal independence can be used to do this:

**Proposition 1** *Let $e$ be a node in a BN and let $c_1, c_2, \ldots, c_m$ be the parents of $e$. If $c_1, c_2, \ldots, c_m$ are causally independent w.r.t. $e$, then the conditional probability $P(e|c_1, \ldots, c_m)$ can be obtained from the conditional probabilities $P(\xi_i|c_i)$ through*

$$P(e = \alpha|c_1, \ldots, c_m) = \sum_{\alpha_1 * \ldots * \alpha_k = \alpha} P(\xi_1 = \alpha_1|c_1) \ldots P(\xi_m = \alpha_m|c_m), \tag{2}$$

*for each value $\alpha$ of $e$. Here $*$ is the base combination operator of $e$.*

**Proof**:[3] The definition of causal independence entails the independence assertions

$$I(\xi_1, \{c_2, \ldots, c_m\}|c_1) \text{ and } I(\xi_1, \xi_2|c_1).$$

By the axiom of weak union (Pearl, 1988, p. 84), we have $I(\xi_1, \xi_2|\{c_1, \ldots, c_m\})$. Thus all of the $\xi_i$ mutually independent given $\{c_1, \ldots, c_m\}$.

Also we have, by the definition of causal independence $I(\xi_1, \{c_2, \ldots, c_m\}|c_1)$, so

$$P(\xi_1|\{c_1, c_2, \ldots, c_m\}) = P(\xi_1|c_1)$$

Thus we have:

$$
\begin{aligned}
P(e = \alpha|c_1, \ldots, c_m) \\
= \quad & P(\xi_1 * \cdots * \xi_m = \alpha|c_1, \ldots, c_m) \\
= \sum_{\alpha_1 * \ldots * \alpha_k = \alpha} & P(\xi_1 = \alpha_1, \ldots, \xi_m = \alpha_m|c_1, \ldots, c_m) \\
= \sum_{\alpha_1 * \ldots * \alpha_k = \alpha} & P(\xi_1 = \alpha_1|c_1, \ldots, c_m) P(\xi_2 = \alpha_2|c_1, \ldots, c_m) \cdots P(\xi_m = \alpha_m|c_1, \ldots, c_m) \\
= \sum_{\alpha_1 * \ldots * \alpha_k = \alpha} & P(\xi_1 = \alpha_1|c_1) P(\xi_2 = \alpha_2|c_2) \cdots P(\xi_m = \alpha_m|c_m) \\
\square
\end{aligned}
$$

The next four sections develop an algorithm for exploiting causal independence in inference.

---

3. Thanks to an anonymous reviewer for helping us to simplify this proof.

## 5. Causal Independence and Heterogeneous Factorizations

In this section, we shall first introduce an operator for combining factors that contain convergent variables. The operator is a basic ingredient of the algorithm to be developed in the next three sections. Using the operator, we shall rewrite equation (2) in a form that is more convenient to use in inference and introduce the concept of heterogeneous factorization.

Consider two factors $f$ and $g$. Let $e_1, ..., e_k$ be the convergent variables that appear in both $f$ and $g$, let $A$ be the list of regular variables that appear in both $f$ and $g$, let $B$ be the list of variables that appear only in $f$, and let $C$ be the list of variables that appear only in $g$. Both $B$ and $C$ can contain convergent variables, as well as regular variables. Suppose $*_i$ is the base combination operator of $e_i$. Then, the *combination $f \otimes g$ of $f$ and $g$* is a function of variables $e_1, ..., e_k$ and of the variables in $A$, $B$, and $C$. It is defined by:[4]

$$
\begin{aligned}
& f \otimes g(e_1 = \alpha_1, \ldots, e_k = \alpha_k, A, B, C) \\
& = \sum_{\alpha_{11} *_1 \alpha_{12} = \alpha_1} \cdots \sum_{\alpha_{k1} *_k \alpha_{k2} = \alpha_k} f(e_1 = \alpha_{11}, \ldots, e_k = \alpha_{k1}, A, B) \\
& \hspace{6cm} g(e_1 = \alpha_{12}, \ldots, e_k = \alpha_{k2}, A, C),
\end{aligned} \tag{3}
$$

for each value $\alpha_i$ of $e_i$. We shall sometimes write $f \otimes g$ as $f(e_1, \ldots, e_k, A, B) \otimes g(e_1, \ldots, e_k, A, C)$ to make explicit the arguments of $f$ and $g$.

Note that base combination operators of different convergent variables can be different.

The following proposition exhibits some of the basic properties of the combination operator $\otimes$.

**Proposition 2** *1. If $f$ and $g$ do not share any convergent variables, then $f \otimes g$ is simply the multiplication of $f$ and $g$. 2. The operator $\otimes$ is commutative and associative.*

**Proof**: The first item is obvious. The commutativity of $\otimes$ follows readily from the commutativity of multiplication and the base combination operators. We shall prove the associativity of $\otimes$ in a special case. The general case can be proved by following the same line of reasoning.

Suppose $f$, $g$, and $h$ are three factors that contain only one variable $e$ and the variable is convergent. We need to show that $(f \otimes g) \otimes h = f \otimes (g \otimes h)$. Let $*$ be the base combination operator of $e$. By the associativity of $*$, we have, for any value $\alpha$ of $e$, that

$$
\begin{aligned}
(f \otimes g) \otimes h(e = \alpha) &= \sum_{\alpha_4 * \alpha_3 = \alpha} f \otimes g(e = \alpha_4) h(e = \alpha_3) \\
&= \sum_{\alpha_4 * \alpha_3 = \alpha} [\sum_{\alpha_1 * \alpha_2 = \alpha_4} f(e = \alpha_1) g(e = \alpha_2)] h(e = \alpha_3) \\
&= \sum_{\alpha_1 * \alpha_2 * \alpha_3 = \alpha} f(e = \alpha_1) g(e = \alpha_2) h(e = \alpha_3) \\
&= \sum_{\alpha_1 * \alpha_4 = \alpha} f(e = \alpha_1) [\sum_{\alpha_2 * \alpha_3 = \alpha_4} g(e = \alpha_2) h(e = \alpha_3)]
\end{aligned}
$$

---

4. Note that the base combination operators under the summations are indexed. With each convergent variable is an associated operator, and we always use the binary operator that is associated with the corresponding convergent variable. In the examples, for ease of exposition, we will use one base combination operator. Where there is more than one type of base combination operator (e.g., we may use 'or', 'sum' and 'max' for different variables in the same network), we have to keep track of which operators are associated with which convergent variables. This will, however, complicate the description.

$$= \sum_{\alpha_1 * \alpha_4 = \alpha} f(e = \alpha_1) g \otimes h(e = \alpha_4)$$

$$= f \otimes (g \otimes h)(e = \alpha).$$

The proposition is hence proved.$\square$

The following propositions give some properties for $\otimes$ that correspond to the operations that we exploited for the algorithm VE. The proofs are straight forward and are omitted.

**Proposition 3** *Suppose $f$ and $g$ are factors and variable $z$ appears in $f$ and not in $g$, then*

$$\sum_z (fg) = (\sum_z f) g, \text{ and}$$

$$\sum_z (f \otimes g) = (\sum_z f) \otimes g.$$

**Proposition 4** *Suppose $f$, $g$ and $h$ are factors such that $g$ and $h$ do not share any convergent variables, then*

$$g(f \otimes h) = (gf) \otimes h. \tag{4}$$

### 5.1 Rewriting Equation 2

Noticing that the contribution variable $\xi_i$ has the same possible values as $e$, we define functions $f_i(e, c_i)$ by

$$f_i(e = \alpha, c_i) = P(\xi_i = \alpha | c_i),$$

for any value $\alpha$ of $e$. We shall refer to $f_i$ as the *contributing factor of $c_i$ to $e$*.

By using the operator $\otimes$, we can now rewrite equation (2) as follows

$$P(e | c_1, \ldots, c_m) = \otimes_{i=1}^m f_i(e, c_i). \tag{5}$$

It is interesting to notice the similarity between equation (1) and equation (5). In equation (1) conditional independence allows one to factorize a joint probability into factors that involve less variables, while in equation (5) causal independence allows one to factorize a conditional probability into factors that involve less variables. However, the ways by which the factors are combined are different in the two equations.

### 5.2 Heterogeneous Factorizations

Consider the Bayesian network in Figure 1. It factorizes the joint probability $P(a, b, c, e_1, e_2, e_3)$ into the following list of factors:

$$P(a), P(b), P(c), P(e_1 | a, b, c), P(e_2 | a, b, c), P(e_3 | e_1, e_2).$$

We say that this factorization is *homogeneous* because all the factors are combined in the same way, i.e., by multiplication.

Now suppose the $e_i$'s are convergent variables. Then their conditional probabilities can be further factorized as follows:

$$P(e_1 | a, b, c) = f_{11}(e_1, a) \otimes f_{12}(e_1, b) \otimes f_{13}(e_1, c),$$
$$P(e_2 | a, b, c) = f_{21}(e_2, a) \otimes f_{22}(e_2, b) \otimes f_{23}(e_2, c),$$
$$P(e_3 | e_1, e_2) = f_{31}(e_3, e_1) \otimes f_{32}(e_3, e_2),$$

where the factor $f_{11}(e_1, a)$, for instance, is the contributing factor of $a$ to $e_1$.

We say that the following list of factors

$$f_{11}(e_1, a), f_{12}(e_1, b), f_{13}(e_1, c), f_{21}(e_2, a), f_{22}(e_2, b), f_{23}(e_2, c), f_{31}(e_3, e_1), f_{32}(e_3, e_2),$$
$$P(a), P(b), \text{ and } P(c) \tag{6}$$

constitute a *heterogeneous factorization* of $P(a, b, c, e_1, e_2, e_3)$ because the joint probability can be obtained by combining those factors in a proper order using either multiplication or the operator $\otimes$. The word heterogeneous is to signify the fact that different factor pairs might be combined in different ways. We call each $f_{ij}$ a *heterogeneous factor* because it needs to be combined with the other $f_{ik}$'s by the operator $\otimes$ before it can be combined with other factors by multiplication. In contrast, we call the factors $P(a)$, $P(b)$, and $P(c)$ *homogeneous factors*.

We shall refer to that heterogeneous factorization as the heterogeneous factorization represented by the BN in Figure 1. It is obvious that this heterogeneous factorization is of finer grain than the homogeneous factorization represented by the BN.

## 6. Flexible Heterogeneous Factorizations and Deputation

This paper extends VE to exploit this finer-grain factorization. We will compute the answer to a query by summing out variables one by one from the factorization just as we did in VE.

The correctness of VE is guaranteed by the fact that factors in a homogeneous factorization can be combined (by multiplication) in any order and by the distributivity of multiplication over summations (see the proof of Theorem 1).

According to Proposition 3, the operator $\otimes$ is distributive over summations. However, factors in a heterogeneous factorization cannot be combined in arbitrary order. For example, consider the heterogeneous factorization (6). While it is correct to combine $f_{11}(e_1, a)$ and $f_{12}(e_1, b)$ using $\otimes$, and to combine $f_{31}(e_3, e_1)$ and $f_{32}(e_3, e_2)$ using $\otimes$, it is not correct to combine $f_{11}(e_1, a)$ and $f_{31}(e_3, e_1)$ with $\otimes$. We want to combine these latter two by multiplication, but only after each has been combined with its sibling heterogeneous factors.

To overcome this difficulty, a transformation called deputation will be performed on our BN. The transformation does not change the answers to queries. And the heterogeneous factorization represented by the transformed BN is flexible in the following sense:

A heterogeneous factorization of a joint probability is *flexible* if:

> The joint probability
> $=$ multiplication of all homogeneous factors
> $\times$ combination (by $\otimes$) of all heterogeneous factors. $\tag{7}$

This property allows us to carry out multiplication of homogeneous factors in arbitrary order, and since $\otimes$ is associative and commutative, combination of heterogeneous factors in arbitrary order. If the conditions of Proposition 4 are satisfied, we can also exchange a multiplication with a combination by $\otimes$. To guarantee the conditions of Proposition 4, the elimination ordering needs to be constrained (Sections 7 and 8).

The heterogeneous factorization of $P(a, b, c, e_1, e_2, e_3)$ given at the end of the previous section is not flexible. Consider combining all the heterogeneous factors. Since the operator $\otimes$ is commutative and associative, one can first combine, for each $i$, all the $f_{ik}$'s, obtaining the conditional probability of $e_i$, and then combine the resulting conditional probabilities. The combination

$$P(e_1|a,b,c) \otimes P(e_2|a,b,c) \otimes P(e_3|e_1,e_2)$$

is not the same as the multiplication

$$P(e_1|a,b,c) P(e_2|a,b,c) P(e_3|e_1,e_2)$$

because the convergent variables $e_1$ and $e_2$ appear in more than one factor. Consequently, equation (7) does not hold and the factorization is not flexible. This problem arises when a convergent variable is shared between two factors that are not siblings. For example, we do not want to combine $f_{11}(e_1,a)$ and $f_{31}(e_3,e_1)$ using $\otimes$. In order to tackle this problem we introduce a new 'deputation' variable so that each heterogeneous factor contains a single convergent variable.

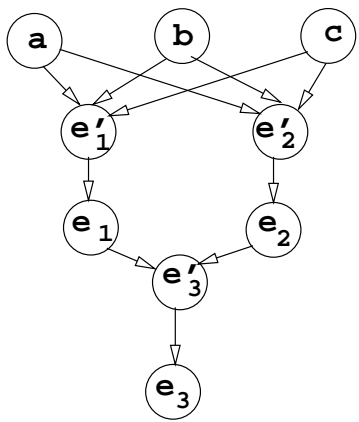

Figure 2: The BN in Figure 1 after the deputation of convergent variables.

Deputation is a transformation that one can apply to a BN to make the heterogeneous factorization represented by the BN flexible. Let $e$ be a convergent variable. To *depute* $e$ is to make a copy $e'$ of $e$, make the parents of $e$ be parents of $e'$, replace $e$ with $e'$ in the contributing factors of $e$, make $e'$ the only parent of $e$, and set the conditional probability $P(e|e')$ as follows:

$$P(e|e') = \begin{cases} 1 & \text{if } e = e' \\ 0 & \text{otherwise} \end{cases} \tag{8}$$

We shall call $e'$ the *deputy* of $e$. The deputy variable $e'$ is a convergent variable by definition. The variable $e$, which is convergent before deputation, becomes a regular variable after deputation. We shall refer to it as a *new regular variable*. In contrast, we shall refer to the variables that are regular before deputation as *old regular variables*. The conditional probability $P(e'|e)$ is a homogeneous factor by definition. It will sometimes be called the *deputing function* and written as $I(e',e)$ since it ensures that $e'$ and $e$ always take the same value.

A *deputation BN* is obtained from a BN by deputing all the convergent variables. In a deputation BN, deputy variables are convergent variables and only deputy variables are convergent variables.

Figure 2 shows the deputation of the BN in Figure 1. It factorizes the joint probability

$$P(a, b, c, e_1, e_1', e_2, e_2', e_3, e_3')$$

into homogeneous factors

$$P(a), P(b), P(c), I_1(e_1', e_1), I_2(e_2', e_2), I_3(e_3', e_3),$$

and heterogeneous factors

$$f_{11}(e_1', a), f_{12}(e_1', b), f_{13}(e_1', c), f_{21}(e_2', a), f_{22}(e_2', b), f_{23}(e_2', c), f_{31}(e_3', e_1), f_{32}(e_3', e_2).$$

This factorization has three important properties.

1. Each heterogeneous factor contains *one and only one* convergent variable. (Recall that the $e_i$'s are no longer convergent variables and their deputies are.)

2. Each convergent variable $e'$ appears in *one and only one* homogeneous factor, namely the deputing function $I(e', e)$.

3. Except for the deputing functions, *none* of the homogeneous factors contain any convergent variables.

Those properties are shared by the factorization represented by any deputation BN.

**Proposition 5** *The heterogeneous factorization represented by a deputation BN is flexible.*

**Proof**: Consider the combination, by $\otimes$, of all the heterogeneous factors in the deputation BN. Since the combination operator $\otimes$ is commutative and associative, we can carry out the combination in following two steps. First for each convergent (deputy) variable $e'$, combine all the heterogeneous factors that contain $e'$, yielding the conditional probability $P(e'|\pi_{e'})$ of $e'$. Then combine those resulting conditional probabilities. It follows from the first property mentioned above that for different convergent variables $e_1'$ and $e_2'$, $P(e_1'|\pi_{e_1'})$ and $P(e_2'|\pi_{e_2'})$ do not share convergent variables. Hence the combination of the $P(e'|\pi_{e'})$'s is just the multiplication of them. Consequently, the combination, by $\otimes$, of all heterogeneous factors in a deputation BN is just the multiplication of the conditional probabilities of all convergent variables. Therefore, we have

> The joint probability of variables in a deputation BN
> = multiplication of conditional probabilities of all variables
> = multiplication of conditional probabilities of all regular variables
> ×multiplication of conditional probabilities of all convergent variables
> = multiplication of all homogeneous factors
> ×combination (by $\otimes$) of all heterogeneous factors.

The proposition is hence proved. □

Deputation does not change the answer to a query. More precisely, we have

**Proposition 6** *The posterior probability $P(X|Y=Y_0)$ is the same in a BN as in its deputation.*

**Proof**: Let $R$, $E$, and $E'$ be the lists of old regular, new regular, and deputy variables in the deputation BN respectively. It suffices to show that $P(R, E)$ is the same in the original BN as in the deputation BN. For any new regular variable $e$, let $e'$ be its deputy. It is easy to see that the quantity $\sum_{e'} I(e', e) P(e'|\pi_{e'})$ in the deputation BN is the same as $P(e|\pi_e)$ in the original BN. Hence,

$$P(R, E) \text{ in the deputation BN}$$
$$= \sum_{E'} P(R, E, E')$$
$$= \sum_{E'} \prod_{r \in R} P(r|\pi_r) \prod_{e \in E} [P(e|\pi_e) P(e'|\pi_{e'})]$$
$$= \prod_{r \in R} P(r|\pi_r) \prod_{e \in E} [\sum_{e'} I(e', e) P(e'|\pi_{e'})]$$
$$= \prod_{r \in R} P(r|\pi_r) \prod_{e \in E} P(e|\pi_e)$$
$$= P(R, E) \text{ in the original BN.}$$

The proposition is proved. □

## 7. Tidy Heterogeneous Factorizations

So far, we have only encountered heterogeneous factorizations that correspond to Bayesian networks. In the following algorithm, the intermediate heterogeneous factorizations do not necessarily correspond by BNs. They do have the property that they combine to form the appropriate marginal probabilities. The general intuition is that the heterogeneous factors must combine with their sibling heterogeneous factors before being multiplied by factors containing the original convergent variable.

In the previous section, we mentioned three properties of the heterogeneous factorization represented by a deputation BN, and we used the first property to show that the factorization is flexible. The other two properties qualify the factorization as a tidy heterogeneous factorization, which is defined below.

Let $z_1$, $z_2$, ..., $z_k$ be a list of variables in a deputation BN such that if a convergent (deputy) variable $e'$ is in $\{z_1, z_2, \ldots, z_k\}$, so is the corresponding new regular variable $e$. A flexible heterogeneous factorization of $P(z_1, z_2, \ldots, z_k)$ is said to be *tidy* If

1. For each convergent (deputy) variable $e' \in \{z_1, z_2, \ldots, z_k\}$, the factorization contains the deputing function $I(e', e)$ and it is the only homogeneous factor that involves $e'$.

2. Except for the deputing functions, none of the homogeneous factors contain any convergent variables.

As stated earlier, the heterogeneous factorization represented by a deputation BN is tidy.

Under certain conditions, to be given in Theorem 3, one can obtain a tidy factorization of $P(z_2, \ldots, z_k)$ by summing out $z_1$ from a tidy factorization of $P(z_1, z_2, \ldots, z_k)$ using the the following procedure.

Procedure `sum-out1`$(\mathcal{F}_1, \mathcal{F}_2, z)$

- Inputs: $\mathcal{F}_1$ — A list of homogeneous factors,
  $\mathcal{F}_2$ — A list of heterogeneous factors,
  $z$ — A variable.

- Output: A list of heterogeneous factors and a list of homogeneous factors.

1. Remove from $\mathcal{F}_1$ all the factors that contain $z$, multiply them resulting in, say, $f$. If there are no such factors, set $f=\texttt{nil}$.

2. Remove from $\mathcal{F}_2$ all the factors that contain $z$, combine them by using $\otimes$ resulting in, say, $g$. If there are no such factors, set $g=\texttt{nil}$.

3. **If** $g=\texttt{nil}$, add the new (homogeneous) factor $\sum_z f$ to $\mathcal{F}_1$.

4. **Else** add the new (heterogeneous) factor $\sum_z fg$ to $\mathcal{F}_2$.

5. Return $(\mathcal{F}_1, \mathcal{F}_2)$.

**Theorem 3** *Suppose a list of homogeneous factors $\mathcal{F}_1$ and a list of heterogeneous factors $\mathcal{F}_2$ constitute a tidy factorization of $P(z_1, z_2, \ldots, z_k)$. If $z_1$ is either a convergent variable, or an old regular variable, or a new regular variable whose deputy is not in the list $\{z_2, \ldots, z_k\}$, then the procedure* $\texttt{sum-out1}(\mathcal{F}_1, \mathcal{F}_2, z_1)$ *returns a tidy heterogeneous factorization of $P(z_2, \ldots, z_k)$.*

The proof of this theorem is quite long and hence is given in the appendix.

## 8. Causal Independence and Inference

Our task is to compute $P(X|Y=Y_0)$ in a BN. According to Proposition 6, we can do this in the deputation of the BN.

An elimination ordering consisting of the variables outside $X \cup Y$ is *legitimate* if each deputy variable $e'$ appears before the corresponding new regular variable $e$. Such an ordering can be found using, with minor adaptations, minimum deficiency search or maximum cardinality search.

The following algorithm computes $P(X|Y=Y_0)$ in the deputation BN. It is called $\text{VE}_1$ because it is an extension of $\text{VE}$.

Procedure $\text{VE}_1(\mathcal{F}_1, \mathcal{F}_2, X, Y, Y_0, \rho)$

- Inputs: $\mathcal{F}_1$ — The list of homogeneous factors
  in the deputation BN;
  $\mathcal{F}_2$ — The list of heterogeneous factors
  in the deputation BN;
  $X$ — A list of query variables;
  $Y$ — A list of observed variables;
  $Y_0$ — The corresponding list of observed values;
  $\rho$ — A legitimate elimination ordering.

- Output: $P(X|Y=Y_0)$.

1. Set the observed variables in all factors to their observed values.

2. **While** $\rho$ is not empty,
   - Remove the first variable $z$ from $\rho$.
   - $(\mathcal{F}_1, \mathcal{F}_2) = \texttt{sum-out1}(\mathcal{F}_1, \mathcal{F}_2, z)$. **Endwhile**

3. Set $h$=multiplication of all factors in $\mathcal{F}_1$
   $\times$ combination (by $\otimes$) of all factors in $\mathcal{F}_2$.
   /* h is a function of variables in $X$. */

4. Return $h(X)/\sum_X h(X)$. /* renormalization */

**Theorem 4** *The output of* $\text{VE}_1(\mathcal{F}_1, \mathcal{F}_2, X, Y, Y_0, \rho)$ *is indeed* $P(X|Y=Y_0)$.

**Proof**: Consider the following modifications to the algorithm. First remove step 1. Then the factor $h$ produced at step 3 is a function of variables in $X$ and $Y$. Add a new step after step 3 that sets the observed variables in $h$ to their observed values. We shall first show that the modifications do not change the output of the algorithm and then show that the output of the modified algorithm is $P(X|Y=Y_0)$.

Let $f(y,-)$, $g(y,-)$, and $h(y,z,-)$ be three functions of $y$ and other variables. It is evident that

$$f(y,-)g(y,-)|_{y=\alpha} = f(y,-)|_{y=\alpha}\, g(y,-)|_{y=\alpha},$$

$$[\sum_z h(y,z,-)]|_{y=\alpha} = \sum_z [h(y,z,-)|_{y=\alpha}].$$

If $y$ is a regular variable, we also have

$$f(y,-)\otimes g(y,-)|_{y=\alpha} = f(y,-)|_{y=\alpha}\otimes g(y,-)|_{y=\alpha}.$$

Consequently, the modifications do not change the output of the procedure.

Since the elimination ordering $\rho$ is legitimate, it is always the case that if a deputy variable $e'$ has not been summed out, neither has the corresponding new regular variable $e$. Let $z_1, ..., z_k$ be the remaining variables in $\rho$ at any time during the execution of the algorithm. Then, $e' \in \{z_1, ..., z_k\}$ implies $e \in \{z_1, ..., z_k\}$. This and the fact that the factorization represented by a deputation BN is tidy enable us to repeatedly apply Theorem 3 and conclude that, after the modifications, the factor created at step 3 is simply the marginal probability $P(X, Y)$. Consequently, the output is $P(X|Y=Y_0)$. $\square$

## 8.1 An Example

This subsection illustrates $\text{VE}_1$ by walking through an example. Consider computing the $P(e_2|e_3=0)$ in the deputation Bayesian network shown in Figure 2. Suppose the elimination ordering $\rho$ is: $a$, $b$, $c$, $e_1'$, $e_2'$, $e_1$, and $e_3'$. After the first step of $\text{VE}_1$,
$\mathcal{F}_1 = \{P(a), P(b), P(c), I_1(e_1', e_1), I_2(e_2', e_2), I_3(e_3', e_3=0)\}$,
$\mathcal{F}_2 = \{f_{11}(e_1', a), f_{12}(e_1', b), f_{13}(e_1', c), f_{21}(e_2', a), f_{22}(e_2', b), f_{23}(e_2', c), f_{31}(e_3', e_1), f_{32}(e_3', e_2)\}$.
Now the procedure enters the while-loop and it sums out the variables in $\rho$ one by one.

After summing out $a$,
$\mathcal{F}_1 = \{P(b), P(c), I_1(e_1', e_1), I_2(e_2', e_2), I_3(e_3', e_3=0)\}$,
$\mathcal{F}_2 = \{f_{12}(e_1', b), f_{13}(e_1', c), f_{22}(e_2', b), f_{23}(e_2', c), f_{31}(e_3', e_1), f_{32}(e_3', e_2), \psi_1(e_1', e_2')\}$,
where $\psi_1(e_1', e_2') = \sum_a P(a)f_{11}(e_1', a)f_{21}(e_2', a)$.

After summing out $b$,
$\mathcal{F}_1 = \{P(c), I_1(e_1', e_1), I_2(e_2', e_2), I_3(e_3', e_3=0)\}$,
$\mathcal{F}_2 = \{f_{13}(e_1', c), f_{23}(e_2', c), f_{31}(e_3', e_1), f_{32}(e_3', e_2), \psi_1(e_1', e_2'), \psi_2(e_1', e_2')\}$,
where $\psi_2(e_1', e_2') = \sum_b P(b)f_{12}(e_1', b)f_{22}(e_2', b)$.

After summing out $c$,

$\mathcal{F}_1 = \{I_1(e_1', e_1), I_2(e_2', e_2), I_3(e_3', e_3{=}0)\}$,

$\mathcal{F}_2 = \{f_{31}(e_3', e_1), f_{32}(e_3', e_2), \psi_1(e_1', e_2'), \psi_2(e_1', e_2'), \psi_3(e_1', e_2')\}$,

where $\psi_3(e_1', e_2') = \sum_c P(c) f_{23}(e_2', c) f_{13}(e_1', c)$.

After summing out $e_1'$,

$\mathcal{F}_1 = \{I_2(e_2', e_2), I_3(e_3', e_3{=}0)\}$,

$\mathcal{F}_2 = \{f_{31}(e_3', e_1), f_{32}(e_3', e_2), \psi_4(e_1, e_2')\}$,

where $\psi_4(e_1, e_2') = \sum_{e_1'} I_1(e_1', e_1)[\psi_1(e_1', e_2') \otimes \psi_2(e_1', e_2') \otimes \psi_3(e_1', e_3')]$.

After summing out $e_2'$,

$\mathcal{F}_1 = \{I_3(e_3', e_3{=}0)\}$,

$\mathcal{F}_2 = \{f_{31}(e_3', e_1), f_{32}(e_3', e_2), \psi_5(e_1, e_2)\}$,

where $\psi_5(e_1, e_2) = \sum_{e_2'} I_2(e_2', e_2) \psi_4(e_1, e_2')$.

After summing out $e_1$,

$\mathcal{F}_1 = \{I_3(e_3', e_3{=}0)\}$,

$\mathcal{F}_2 = \{f_{32}(e_3', e_2), \psi_6(e_3', e_2)\}$,

where $\psi_6(e_3', e_2) = \sum_{e_1} f_{31}(e_3', e_1) \psi_5(e_1, e_2)$.

Finally, after summing out $e_3'$,

$\mathcal{F}_1 = \emptyset$,

$\mathcal{F}_2 = \{\psi_7(e_2)\}$,

where $\psi_7(e_2) = \sum_{e_3'} I_3(e_3', e_3{=}0)[f_{32}(e_3', e_2) \otimes \psi_6(e_3', e_2)]$. Now the procedure enters step 3, where there is nothing to do in this example. Finally, the procedure returns $\psi_7(e_2) / \sum_{e_2} \psi_7(e_2)$, which is $P(e_2|e_3{=}0)$, the required probability.

## 8.2 Comparing VE and VE$_1$

In comparing VE and VE$_1$, we notice that when summing out a variable, they both combine only those factors that contain the variable. However, the factorization that the latter works with is of finer grain than the factorization used by the former. In our running example, the latter works with a factorization which initially consists of factors that contain only two variables; while the factorization the former uses initially include factors that contain five variables. On the other hand, the latter uses the operator $\otimes$ which is more expensive than multiplication. Consider, for instance, calculating $f(e, a) \otimes g(e, b)$. Suppose $e$ is a convergent variable and all variables are binary. Then the operation requires $2^4$ numerical multiplications and $2^4 - 2^3$ numerical summations. On the other hand, multiplying $f(e, a)$ and $g(e, b)$ only requires $2^3$ numerical multiplications.

Despite the expensiveness of the operator $\otimes$, VE$_1$ is more efficient than VE. We shall provide empirical evidence in support of this claim in Section 10. To see a simple example where this is true, consider the BN in Figure 3(1), where $e$ is a convergent variable. Suppose all variables are binary. Then, computing $P(e)$ by VE using the elimination ordering $c_1$, $c_2$, $c_3$, and $c_4$ requires $2^5 + 2^4 + 2^3 + 2^2{=}60$ numerical multiplications and $(2^5 - 2^4) + (2^4 - 2^3) + (2^3 - 2^2) + (2^2 - 2){=}30$ numerical additions. On the other hand, computing $P(e)$ in the deputation BN shown in Figure 3(2) by VE$_1$ using the elimination ordering $c_1$, $c_2$, $c_3$, $c_4$, and $e'$ requires only $2^2 + 2^2 + 2^2 + 2^2 + (3 \times 2^2 + 2^2){=}32$ numerical multiplications and $2 + 2 + 2 + 2 + (3 \times 2 + 2){=}16$ numerical additions. Note that summing out $e'$ requires $3 \times 2^2 + 2^2$ numerical multiplications because after summing out $c_i$'s, there are four heterogeneous factors, each containing the only argument $e'$. Combining them pairwise requires $3 \times 2^2$ multiplications. The resultant factor needs to be multiplied with the deputing factor $I(e', e)$, which requires $2^2$ numerical multiplications.

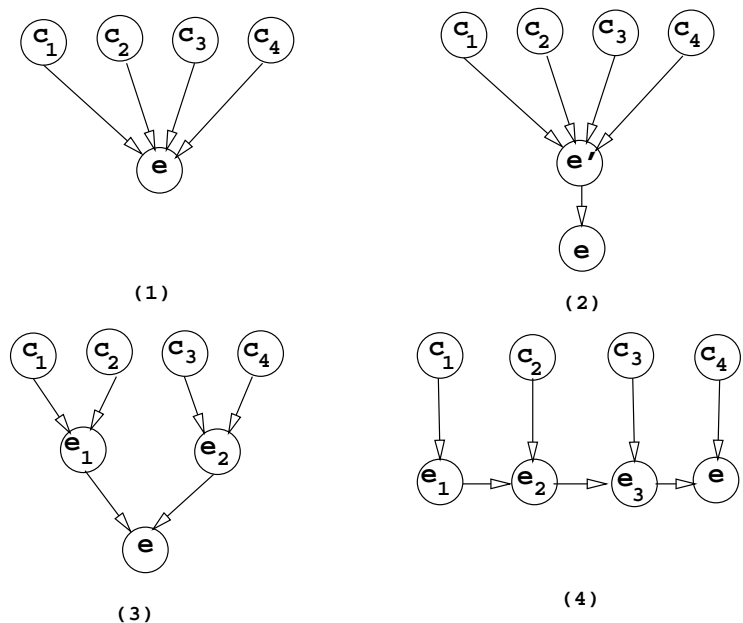

Figure 3: A BN, its deputation and transformations.

## 9. Previous Methods

Two methods have been proposed previously for exploiting causal independence to speed up inference in general BNs (Olesen et al., 1989; Heckerman, 1993). They both use causal independence to transform the topology of a BN. After the transformation, conventional algorithms such as CTP or VE are used for inference.

We shall illustrate those methods by using the BN in Figure 3(1). Let $*$ be the base combination operator of $e$, let $\xi_i$ denote the contribution of $c_i$ to $e$, and let $f_i(e, c_i)$ be the contributing factor of $c_i$ to $e$.

The *parent-divorcing* method (Olesen et al., 1989) transforms the BN into the one in Figure 3(3). After the transformation, all variables are regular and the new variables $e_1$ and $e_2$ have the same possible values as $e$. The conditional probabilities of $e_1$ and $e_2$ are given by

$$P(e_1|c_1, c_2) = f_1(e, c_1) \otimes f_2(e, c_2),$$

$$P(e_2|c_3, c_4) = f_3(e, c_3) \otimes f_4(e, c_4).$$

The conditional probability of $e$ is given by

$$P(e = \alpha | e_1 = \alpha_1, e_2 = \alpha_2) = 1 \text{ if } \alpha = \alpha_1 * \alpha_2,$$

for any value $\alpha$ of $e$, $\alpha_1$ of $e_1$, and $\alpha_2$ of $e_2$. We shall use PD to refer to the algorithm that first performs the parent-divorcing transformation and then uses VE for inference.

The *temporal transformation* by Heckerman (1993) converts the BN into the one in Figure 3(4). Again all variables are regular after the transformation and the newly introduced variables have the same possible values as $e$. The conditional probability of $e_1$ is given by

$$P(e_1 = \alpha | c_1) = f_1(\xi_1 = \alpha, c_1),$$

for each value $\alpha$ of $e_1$. For $i = 2, 3, 4$, the conditional probability of $e_i$ ($e_4$ stands for $e$) is given by

$$P(e_i = \alpha | e_{i-1} = \alpha_1, c_i) = \sum_{\alpha_1 * \alpha_2 = \alpha} f_i(e = \alpha_2, c_i),$$

for each possible value $\alpha$ of $e_i$ and $\alpha_1$ of $e_{i-1}$. We shall use TT to refer to the algorithm that first performs the temporal transformation and then uses VE for inference.

The factorization represented by the original BN includes a factor that contain five variables, while factors in the transformed BNs contain no more than three variables. In general, the transformations lead to finer-grain factorizations of joint probabilities. This is why PD and TT can be more efficient than VE .

However, PD and TT are not as efficient as VE$_1$ . We shall provide empirical evidence in support of this claim in the next section. Here we illustrate it by considering calculating $P(e)$. Doing this in Figure 3(3) by VE using the elimination ordering $c_1$, $c_2$, $c_3$, $c_4$, $e_1$, and $e_2$ would require $2^3 + 2^2 + 2^3 + 2^2 + 2^3 + 2^2 = 36$ numerical multiplications and $18$ numerical additions.[5] Doing the same in Figure 3(4) using the elimination ordering $c_1$, $e_1$, $c_2$, $e_2$, $c_3$, $e_3$, $c_4$ would require $2^2 + 2^3 + 2^2 + 2^3 + 2^2 + 2^3 + 2^2 = 40$ numerical multiplications and $20$ numerical additions. In both cases, more numerical multiplications and additions are performed than VE$_1$ . The differences are more drastic in complex networks, as will be shown in the next section.

The saving for this example may seem marginal. It may be reasonable to conjecture that, as Oleson's method produces families with three elements, this marginal saving is all that we can hope for; producing factors of two elements rather than cliques of three elements. However, interacting causal variables can make the difference more extreme. For example, if we were to use Oleson's method for the BN of Figure 1, we produce[6] the network of Figure 4. Any triangulation for this network has at least one clique with four or more elements, yet VE$_1$ does not produce a factor with more than two elements.

Note that as far as computing $P(e)$ in the networks shown in Figure 3 is concerned, VE$_1$ is more efficient than PD, PD is more efficient than TT, and TT is more efficient than VE. Our experiments show that this is true in general.

---

5. This is exactly the same number of operations required to determine $P(e)$ using clique-tree propagation on the same network. The clique tree for Figure 3(3) has three cliques, one containing $\{c_1, c_2, e_1\}$, one containing $\{c_3, c_4, e_2\}$, and once containing $\{e_1, e_2, e\}$. The first clique contains 8 elements; to construct it requires $2^2 + 2^3 = 12$ multiplications. The message that needs to be sent to the third clique is the marginal on $e_1$ which is obtained by summing out $c_1$ and $c_2$. Similarly for the second clique. The third clique again has 8 elements and requires $12$ multiplications to construct. In order to extract $P(e)$ from this clique, we need to sum out $e_1$ and $e_2$. This shown one reason why VE$_1$ can be more efficient that CTP or VE; VE$_1$ never constructs a factor with three variables for this example. Note however, that an advantage of CTP is that the cost of building the cliques can be amortized over many queries.

6. Note that we need to produce two variables both of which represent "noisy" $a * b$. We need two variables as the noise applied in each case is independent. Note that if there was no noise in the network — if $e_1 = a * b * c$ — we only need to create one variable, but also $e_1$ and $e_2$ would be the same variable (or at least be perfectly correlated). In this case we would need a more complicated example to show our point.

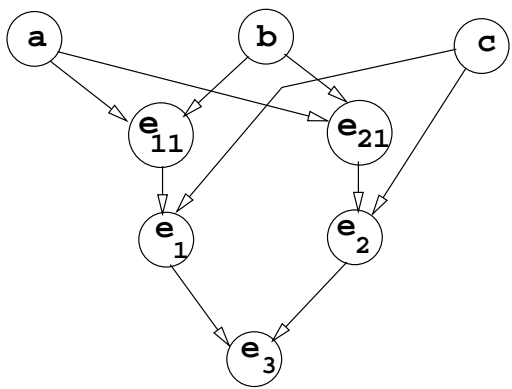

Figure 4: The result of Applying Oleson's method to the BN of Figure 1.

## 10. Experiments

The CPCS networks are multi-level, multi-valued BNs for medicine. They were created by Pradhan *et al.* (1994) based on the Computer-based Patient Case Simulation system (CPCS-PM) developed by Parker and Miller (1987). Two CPCS networks[7] were used in our experiments. One of them consists of 422 nodes and 867 arcs, and the other contains 364 nodes. They are among the largest BNs in use at the present time.

The CPCS networks contain abundant causal independencies. As a matter of fact, each non-root variable is a convergent variable with base combination operator MAX. They are good test cases for inference algorithms that exploit causal independencies.

### 10.1 CTP-based Approaches versus VE-based Approaches

As we have seen in the previous section, one kind of approach for exploiting causal independencies is to use them to transform BNs. Thereafter, any inference algorithms, including CTP or VE, can be used for inference.

We found the coupling of the network transformation techniques and CTP was not able to carry out inference in the two CPCS networks used in our experiments. The computer ran out memory when constructing clique trees for the transformed networks. As will be reported in the next subsection, however, the combination of the network transformation techniques and VE was able to answer many queries.

This paper has proposed a new method of exploiting causal independencies. We have observed that causal independencies lead to a factorization of a joint probability that is of finer-grain than the factorization entailed by conditional independencies alone. One can extend any inference algorithms, including CTP and VE, to exploit this finer-grain factorization. This paper has extended VE and obtained an algorithm called $VE_1$. $VE_1$ was able to answer almost all queries in the two CPCS networks. We conjecture, however, that an extension of CTP would not be able to carry out inference with the two CPCS networks at all. Because the resources that $VE_1$ takes to answer any query in a BN can be no more than those an extension of CTP would take to construct a clique tree for the BN and there are, as will be seen in the next subsection, queries in the two CPCS networks that $ve_1$ was not able to answer.

---

7. Obtained from `ftp://camis.stanford.edu/pub/pradhan`. The file names are `CPCS-LM-SM-K0-V1.0.txt` and `CPCS-networks/std1.08.5`.

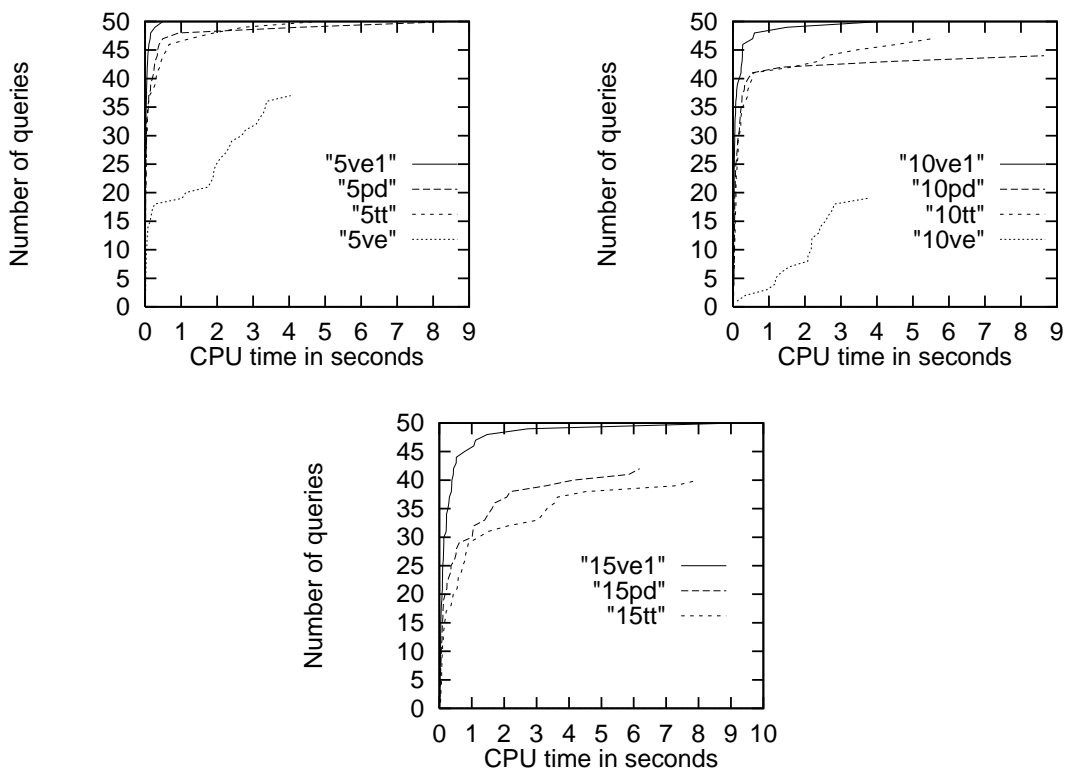

Figure 5: Comparisons in the 364-node BN.

In summary, CTP based approaches are not or would not be able to deal with the two CPCS networks, while $ve$-based approaches can (to different extents).

## 10.2 Comparisons of $ve$-based Approaches

This subsection provides experimental data to compare the $ve$-based approaches namely PD, TT, and $ve_1$. We also compare those approaches with $ve$ itself to determine how much can be gained by exploiting causal independencies.

In the 364-node network, three types of queries with one query variable and five, ten, or fifteen observations respectively were considered. Fifty queries were randomly generated for each query type. A query is passed to the algorithms after nodes that are irrelevant to it have been pruned. In general, more observations mean less irrelevant nodes and hence greater difficulty to answer the query. The CPU times the algorithms spent in answering those queries were recorded.

In order to get statistics for all algorithms, CPU time consumption was limited to ten seconds and memory consumption was limited to ten megabytes.

The statistics are shown in Figure 5. In the charts, the curve "5ve1", for instance, displays the time statistics for $ve_1$ on queries with five observations. Points on the X-axis represent CPU times in seconds. For any time point, the corresponding point on the Y-axis represents the number of five-observation queries that were *each* answered within the time by $\text{VE}_1$.

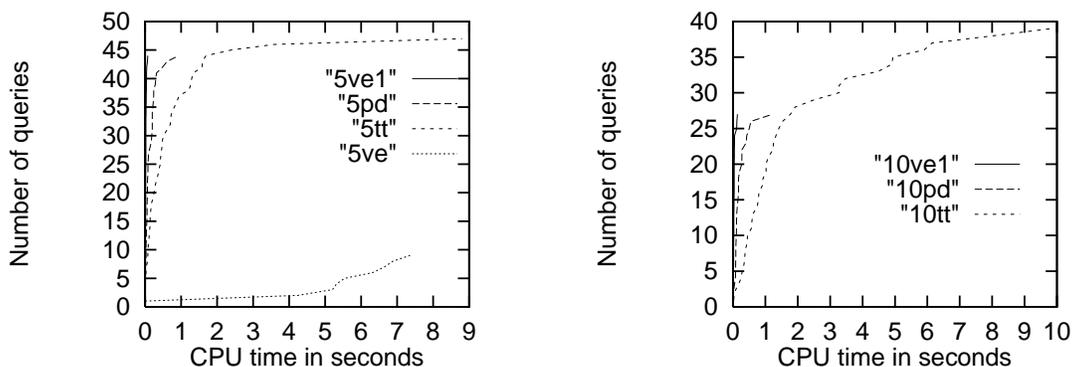

Figure 6: Comparisons in the 422-node BN.

We see that while $\text{VE}_1$ was able to answer all the queries, PD and TT were not able to answer some of the ten-observation and fifteen-observation queries. VE was not able to answer a majority of the queries.

To get a feeling about the average performances of the algorithms, regard the curves as representing functions of $y$, instead of $x$. The integration, along the $Y$-axis, of the curve "10PD", for instance, is roughly the total amount of time PD took to answer all the ten-observation queries that PD was able to answer. Dividing this by the total number of queries answered, one gets the average time PD took to answer a ten-observation query.

It is clear that on average, $\text{VE}_1$ performed significantly better than PD and TT, which in turn performed much better than VE. The average performance of PD on five- or ten-observation queries are roughly the same as that of TT, and slightly better on fifteen-observation queries.

In the 422-node network, two types of queries with five or ten observations were considered and fifty queries were generated for each type. The same space and time limits were imposed as in the 364-node networks. Moreover, approximations had to be made; real numbers smaller than 0.00001 were regarded as zero. Since the approximations are the same for all algorithms, the comparisons are fair.

The statistics are shown in Figure 6. The curves "5ve1" and "10ve1" are hardly visible because they are very close to the $Y$-axis.

Again we see that on average, $\text{VE}_1$ performed significantly better than PD, PD performed significantly better than TT, and TT performed much better than VE.

One might notice that TT was able to answer thirty nine ten-observation queries, more than that $\text{VE}_1$ and PD were able to. This is due to the limit on memory consumption. As we will see in the next subsection, with the memory consumption limit increased to twenty megabytes, $\text{VE}_1$ was able to answer forty five ten-observation queries *exactly* under ten seconds.

## 10.3 Effectiveness of $\text{VE}_1$

We have now established that $\text{VE}_1$ is the most efficient VE-based algorithm for exploiting causal independencies. In this section we investigate how effective $\text{VE}_1$ is.

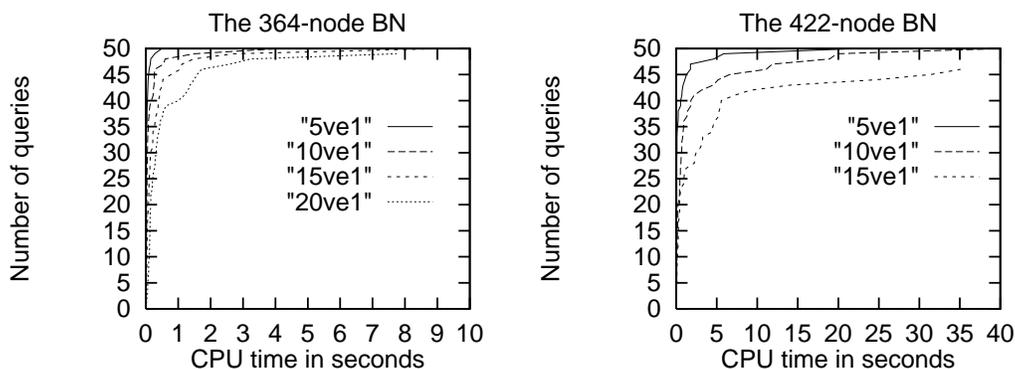

Figure 7: Time statistics for $\text{VE}_1$.

Experiments have been carried out in both of the two CPCS networks to answer this question. In the 364-node network, four types of queries with one query variable and five, ten, fifteen, or twenty observations respectively were considered. Fifty queries were randomly generated for each query type. The statistics of the times $\text{VE}_1$ took to answer those queries are given by the left chart in Figure 7. When collecting the statistics, a ten MB memory limit and a ten second CPU time limit were imposed to guide against excessive resource demands. We see that all fifty five-observation queries in the network were each answered in less than half a second. Forty eight ten-observation queries, forty five fifteen-observation queries, and forty twenty-observation queries were answered in one second. There is, however, one twenty-observation query that $\text{VE}_1$ was not able to answer within the time and memory limits.

In the 364-node network, three types of queries with one query variable and five, ten, or fifteen, observations respectively were considered. Fifty queries were randomly generated for each query type. Unlike in the previous section, *no approximations* were made. A twenty MB memory limit and a forty-second CPU time limit were imposed. The time statistics is shown in the right hand side chart. We see that $\text{VE}_1$ was able to answer all most all queries and a majority of the queries were answered in little time. There are, however, three fifteen-observation queries that $\text{VE}_1$ was not able to answer.

## 11. Conclusions

This paper has been concerned with how to exploit causal independence in exact BN inference. Previous approaches (Olesen et al., 1989; Heckerman, 1993) use causal independencies to transform BNs. Efficiency is gained because inference is easier in the transformed BNs than in the original BNs.

A new method has been proposed in this paper. Here is the basic idea. A Bayesian network can be viewed as representing a factorization of a joint probability into the multiplication of a list of conditional probabilities. We have studied a notion of causal independence that enables one to further factorize the conditional probabilities into a combination of even smaller factors and consequently obtain a finer-grain factorization of the joint probability.

We propose to extend inference algorithms to make use of this finer-grain factorization. This paper has extended an algorithm called VE. Experiments have shown that the extended VE algo-rithm, $\text{VE}_1$, is significantly more efficient than if one first performs Olesen *et al.*'s or Heckerman's transformation and then apply $\text{VE}$.

The choice of $\text{VE}$ instead of the more widely known CTP algorithm is due to its ability to work in networks that CTP cannot deal with. As a matter of fact, CTP was not able to deal with the networks used in our experiments, even after Olesen *et al.*'s or Heckerman's transformation. On the other hand, $\text{VE}_1$ was able to answer almost all randomly generated queries with and a majority of the queries were answered in little time. It would be interesting to extend CTP to make use of the finer-grain factorization mentioned above.

As we have seen in the previous section, there are queries, especially in the 422-node network, that took $\text{VE}_1$ a long time to answer. There are also queries that $\text{VE}_1$ was not able to answer. For those queries, approximation is a must. We employed an approximation technique when comparing algorithms in the 422-node network. The technique captures, to some extent, the heuristic of ignoring minor distinctions. In future work, we are developing a way to bound the error of the technique and an anytime algorithm based on the technique.

## Acknowledgements

We are grateful to Malcolm Pradhan and Gregory Provan for sharing with us the CPCS networks. We also thank Jack Breese, Bruce D'Ambrosio, Mike Horsch, Runping Qi, and Glenn Shafer for valuable discussions, and Ronen Brafman, Chris Geib, Mike Horsch and the anonymous reviewers for very helpful comments. Mr. Tak Yin Chan has been a great help in the experimentations. Research was supported by NSERC Grant OGPOO44121, the Institute for Robotics and Intelligent Systems, Hong Kong Research Council grant HKUST658/95E and Sino Software Research Center grant SSRC95/96.EG01.

## Appendix A. Proof of Theorem 3

**Theorem 3** Suppose a list of homogeneous factors $\mathcal{F}_1$ and a list of heterogeneous factors $\mathcal{F}_2$ constitute a tidy factorization of $P(z_1, z_2, \ldots, z_k)$. If $z_1$ is either a convergent variable, or an old regular variable, or a new regular variable whose deputy is not in the list $\{z_2, \ldots, z_k\}$, then the procedure `sum-out1`$(\mathcal{F}_1, \mathcal{F}_2, z_1)$ returns a tidy heterogeneous factorization of $P(z_2, \ldots, z_k)$.

**Proof**: Suppose $f_1, \ldots, f_r$ are all the heterogeneous factors and $g_1, \ldots, g_s$ are all the homogeneous factors. Also suppose $f_1, \ldots, f_l, g_1, \ldots, g_m$ are all the factors that contain $z_1$. Then

$$
\begin{aligned}
P(z_2, \ldots, z_k) &= \sum_{z_1} P(z_1, z_2, \ldots, z_k) \\
&= \sum_{z_1} \otimes_{j=1}^r f_j \prod_{i=1}^s g_i \\
&= \sum_{z_1} [(\otimes_{j=1}^l f_j) \otimes (\otimes_{j=l+1}^r f_j)] \prod_{i=1}^m g_i \prod_{i=m+1}^s g_i \\
&= \sum_{z_1} [(\otimes_{j=1}^l f_j \prod_{i=1}^m g_i) \otimes (\otimes_{j=l+1}^r f_j)] \prod_{i=m+1}^s g_i \qquad (9)
\end{aligned}
$$

$$= [(\sum_{z_1} \otimes_{j=1}^l f_j \prod_{i=1}^m g_i) \otimes (\otimes_{j=l+1}^r f_j)] \prod_{i=m+1}^s g_i, \qquad (10)$$

where equation (10) is due to Proposition 3. Equation (9) is follows from Proposition 4. As a matter of fact, if $z_1$ is a convergent variable, then it is the only convergent variable in $\prod_{i=1}^m g_i$ due to the first condition of tidiness. The condition of Proposition 4 is satisfied because $z_1$ does not appear in $f_{l+1}, ..., f_r$. On the other hand, if $z_1$ is an old regular variable or a new regular variable whose deputy does not appear in the list $z_2, ..., z_k$, then $\prod_{i=1}^m g_i$ contains no convergent variables due to the second condition of tidiness. Again the condition of Proposition 4 is satisfied. We have thus proved that `sum-out1`$(\mathcal{F}_1, \mathcal{F}_2, z_1)$ yields a flexible heterogeneous factorization of $P(z_2, ..., z_k)$.

Let $e'$ be a convergent variable in the list $z_2, ..., z_k$. Then $z_1$ cannot be the corresponding new regular variable $e$. Hence the factor $I(e', e)$ is not touched by `sum-out1`$(\mathcal{F}_1, \mathcal{F}_2, z_1)$. Consequently, if we can show that the new factor created by `sum-out1`$(\mathcal{F}_1, \mathcal{F}_2, z_1)$ is either a heterogeneous factor or a homogeneous factor that contain no convergent variable, then the factorization returned is tidy.

Suppose `sum-out1`$(\mathcal{F}_1, \mathcal{F}_2, z_1)$ does not create a new homogeneous factor. Then no heterogeneous factors in $\mathcal{F}_1$ contain $z_1$. If $z_1$ is a convergent variable, say $e'$, then $I(e', e)$ is the only homogeneous factor that contain $e'$. The new factor is $\sum_{e'} I(e', e)$, which does contain any convergent variables. If $z_1$ is an old regular variable or a new regular variable whose deputy is not in the list $z_2, ..., z_k$, all the factors that contain $z_1$ do not contain any convergent variables. Hence the new factor again does not contain any convergent variables. The theorem is thus proved. □